\documentclass[times,twocolumn,final]{elsarticle}

\usepackage{cag}
\usepackage{framed,multirow}

\usepackage{amssymb}
\usepackage{amsmath,amssymb,amsfonts}%
\usepackage{mathrsfs}%
\usepackage{amsthm}%
\usepackage{algorithm}%
\usepackage{algorithmicx}
\usepackage{algpseudocode}%

\usepackage{url}
\usepackage{xcolor}
\definecolor{newcolor}{rgb}{.8,.349,.1}

\usepackage{hyperref}

\usepackage[switch,pagewise]{lineno} 

\journal{Computers \& Graphics}

\begin{document}

\verso{Preprint Submitted for review}

\begin{frontmatter}

\title{Graph Transformer for 3D point clouds classification and semantic segmentation}

\author[1]{Wei \snm{Zhou}\corref{cor1}\fnref{fn1}}
\cortext[cor1]{Corresponding author;}
\emailauthor{mczhouwei12@gmail.com}{Wei Zhou}
\emailauthor{qianwang7961@gmail.com}{Qian Wang}
\emailauthor{weiweijin1109@gmail.com}{Weiwei Jin}
\emailauthor{xinzheshi1@gmail.com}{Xinzhe Shi}
\emailauthor{yhe@ntu.edu.sg}{Ying He}
\author[1]{Qian \snm{Wang}\fnref{fn1}}
\author[1]{Weiwei \snm{Jin}}
\author[1]{Xinzhe \snm{Shi}}
\author[2]{Ying \snm{He}}
\fntext[fn1]{These authors contributed equally to this work.} 
\address[1]{School of Information Science and Technology, Northwest University, Xi'an, 710127, China}
\address[2]{School of Computer Science and Engineering, Nanyang Technological University, 639798, Singapore}

\received{\today}

\begin{abstract}
Recently, graph-based and Transformer-based deep learning have demonstrated excellent performances on various point cloud tasks. 
Most of the existing graph-based methods rely on static graph, which take a fixed input to establish graph relations. Moreover, many graph-based methods apply maximizing and averaging to aggregate neighboring features, so that only a single neighboring point affects the feature of centroid or different neighboring points own the same influence on the centroid's feature, which ignoring the correlation and difference between points. 
Most Transformer-based approaches extract point cloud features based on global attention and lack the feature learning on local neighbors. 
To solve the above issues of graph-based and Transformer-based models, we propose a new feature extraction block named Graph Transformer and construct a 3D point cloud learning network called GTNet to learn features of point clouds on local and global patterns. Graph Transformer integrates the advantages of graph-based and Transformer-based methods, and consists of Local Transformer that use intra-domain cross-attention and Global Transformer that use global self-attention.
Finally, we use GTNet for shape classification, part segmentation and semantic segmentation tasks in this paper. The experimental results show that our model achieves good learning and prediction ability on most tasks. The source code and pre-trained model of GTNet will be released on \href{https://github.com/NWUzhouwei/GTNet}{https://github.com/NWUzhouwei/GTNet}.
\end{abstract}


\end{frontmatter}


\section{Introduction}\label{sec:Introduction}

\begin{figure*}[htbp]
	\centering
 \setlength{\abovecaptionskip}{0.cm} 
\setlength{\abovecaptionskip}{0.cm} 
\setlength{\belowdisplayskip}{3pt} 	
	\includegraphics[width=1\textwidth]{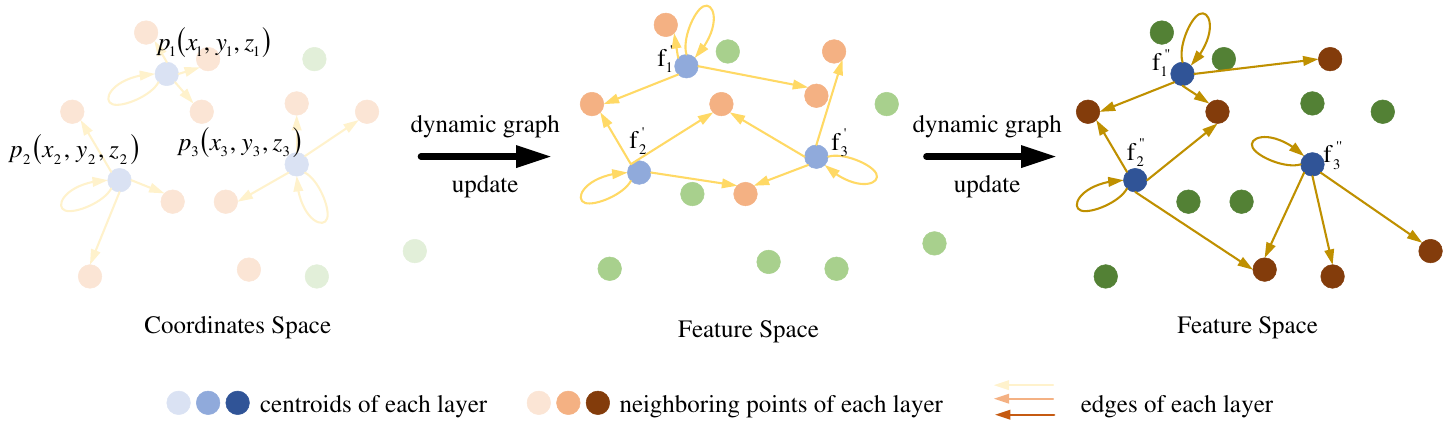}
	\caption{Updating process of dynamic graph in coordinate space and feature spaces. The figure shows the dynamic graph establishment of three centroids, the neighboring points $K$ of centroids are set to 4 when performing K-NN, $p_{i}$ ($i=1,2,3$) are the coordinates of points, $f^{'}_{i}$ and $f^{''}_{i}$ are the features of points, where $f^{''}_{i}$ is the deep feature of $f^{'}_{i}$.}
	\label{fig:update}
\end{figure*}

Recently, many researchers have achieved significant results~\cite{i2pmae,pvcnn,xu2024probability,zheng2023point} in applying deep learning to 3D point clouds.In point cloud data, there may be correlations between points, similarities between features of adjacent points, and differences between features of distant points. Many graph-based methods have been designed and proposed to take full advantage of this property~\cite{wu2024mitigating,zhang2023improving,zhou2023multi}. 
Graph-based methods can be roughly divided into static and dynamic graph~\cite{graph}. Due to the greater flexibility of dynamic graph compared to static graph and it ability to better convey deep feature, dynamic graph is more suitable for point cloud learning.
However, the design of dynamic graph structure is more complicated, and it is necessary to consider when and how to establish the graph dependencies. Another issue is which aggregation method to take among the neighboring points to obtain the features of centroid after the edges of the graph are established. Most of the existing methods adopt max-pooling to directly select a unique neighboring features as the features of centroid, or use the same weight to sum all neighbor point features to obtain centroid features. However, in the feature graph, the dependencies between different neighboring points and centroids are different~\cite{dgcnn}, so different weights should be assigned to each neighboring points.

In the fields of natural language processing (NLP) and image analysis, Transformer has achieved great results~\cite{attention,image}. Recently, many methods have designed Transformer-based deep learning models for point clouds and achieved good performances~\cite{pt,cloudtf,wu2023towards}. The self-attention mechanism takes into account the sequence invariance of the irregular input data, which shows the high fitness between the self-attention mechanism and point clouds. 
However, most existing methods apply Transformers only to global areas, often neglecting local feature extraction. Yet, local information is crucial for effective point cloud learning.

In this paper, we discovered that the fusion of graph-based and Transformer-based methods can reasonably solve their respective problems. The graph-based approach can well obtain the dependencies between points on local neighborhoods; the Transformer-based method can assign different weights to each neighboring points and learn global deep features. 
Thus, we propose a new deep learning network GTNet to process point cloud data by using Encoder-Decoder structure, which combines the advantages of graph-based and Transformer-based approaches. 
It is worth stating that to reduce the loss of features due to downsampling, we view all input points as centroids in GTNet. 
GTNet is mainly composed of feature extraction blocks (Graph Transformer), which are mainly divided into two sub modules: Local Transformer and Global Transformer.
In Local Transformer, as shown in Figure~\ref{fig:update}, we firstly build a dynamic graph to generate the edges between the centroids and neighbors by the current input, then calculate distinct weights for each point by the intra-domain cross-attention (inter-domain cross-attention is a type of vector attention that performs subtraction operations between query vectors and key vectors to generate attention weights), and conduct weight summation of features for different neighboring points which have edge relations, thus to obtain the local features.  
In Global Transformer, we use the global self-attention to generate new centroid features based on the attention weights of all centroids. This process can achieve more contextual representation of points than Local Transformer. 

The model GTNet designed in this paper can be used to handle a variety of point cloud tasks. We adopt ModelNet40 dataset for classification experiments, achieving 93.2\% OA and 92.6\% mAcc; we implement part segmentation on the ShapeNet Part dataset with the evaluation metric mIoU of 85.1\%; we also carry out semantic segmentation tasks on the S3DIS dataset with the evaluation metric mIoU of 64.3\%.
The main contributions of the paper are as follows:

$\bullet$ We propose a deep learning model GTNet which is based on the fusion of dynamic graph and Transformer.

$\bullet$ We design a two sub-structures of the feature extraction block named Graph Transformer to extract point cloud features.

$\bullet$ We design dynamic graphs in the feature extraction layer to obtain deep feature, and add feature encoding in the local transformer to enhance the perception of local shapes.

$\bullet$ We apply GTNet on ModelNet40, ShapeNet Part, and S3DIS datasets. The experimental results illustrate that GTNet can achieve good classification and segmentation metrics.
\section{Related work}
\label{sec:Related_work}
\begin{figure*}[htbp]
	\centering
        \setlength{\abovecaptionskip}{0.cm} 
        \setlength{\abovecaptionskip}{0.cm} 
        \setlength{\belowdisplayskip}{3pt}
	\includegraphics[width=0.99\textwidth]{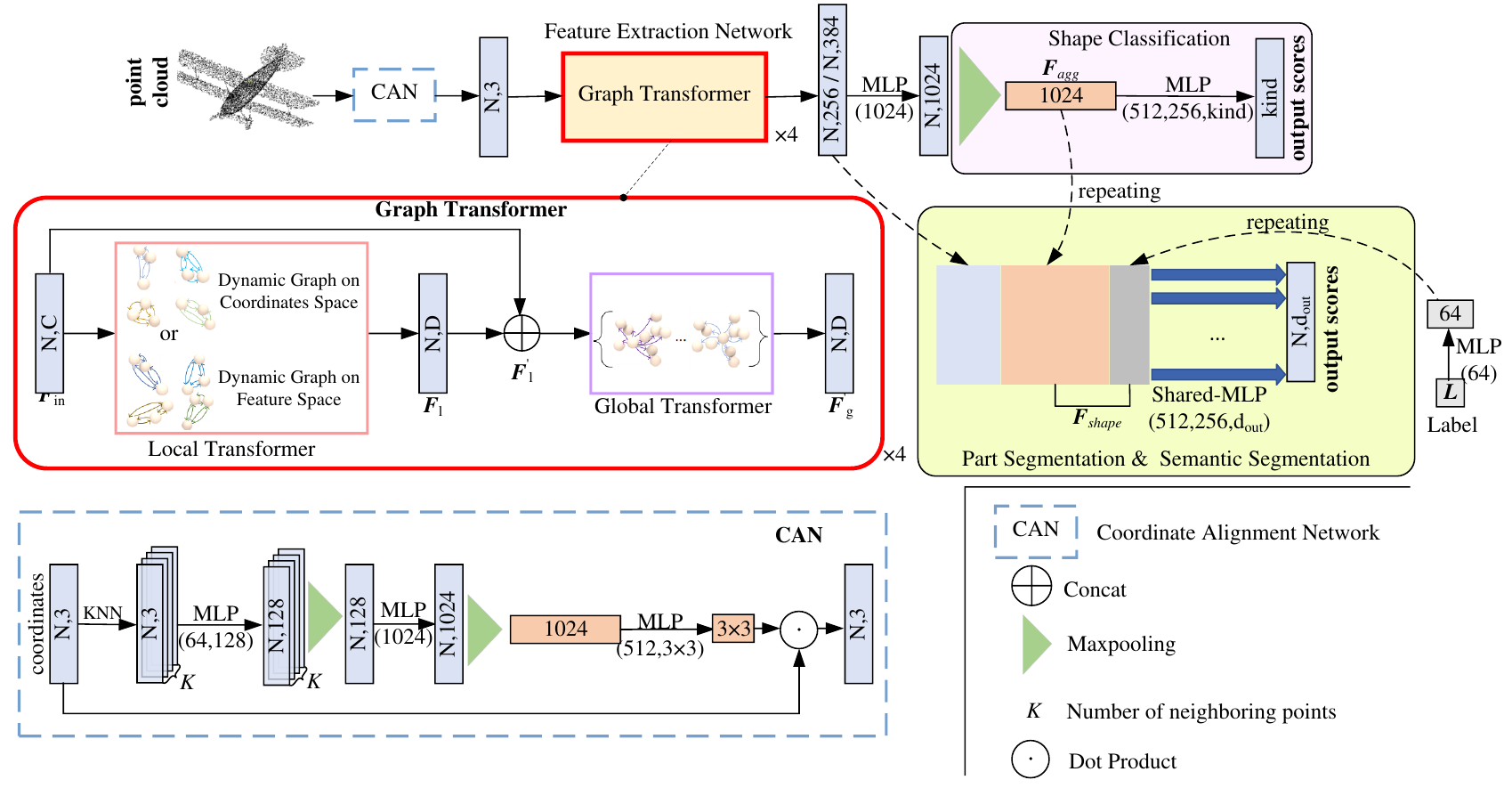}
	\caption{The specific architecture diagram of Graph Transformer. The feature extraction network consists of four feature extraction blocks named Graph Transformer, which is composed of Local Transformer and Global Transformer. The Local Transformer uses the intra-domain cross-attention mechanism based on the dynamic graph structure to obtain local features of the point clouds, and the Global Transformer uses the global self-attention mechanism to obtain global features of the point clouds. In shape classification task, the output dimensions of each feature extraction layer are 64, 64,128,256, respectively. In part segmentation and semantic segmentation tasks, the output dimension of each feature extraction layer is 96, and after the last feature extraction layer, the outputs of the four layers are concatenated to obtain 384 dimensional features. Coordinate Adjustment Network is used to enhance the invariance of rotation and translation.}
	\label{fig:architecture}
\end{figure*}
\textbf{Graph-based model.} Graph-based methods utilize the geometric relations between points to establish dependencies, and aggregate neighboring information to obtain the features of centroids, where the centroids are regarded as the vertexes of the graph, and the dependencies between the centroids and neighboring points are considered as the directed edges. Graph-based methods fall into two categories: static~\cite{tgnet,novel,oversegmentation} and dynamic~\cite{dgcnn,kcnet,dpam,meteornet,sat,dynamic}. The static graph-based methods use a graph consisting of fixed vertexes and edges in each layer of the model for deep learning, and most existing methods use this structure with simplicity and low time consumption~\cite{tgnet,oversegmentation}.
The dynamic graph-based  approaches~\cite{dgcnn,dpam,meteornet,sat,chen2023ddgcn} dynamically update the graph structure by the output features of each layer, thus could adjust and optimize the point features according to the other points.
Rozza et al. proposed a graph-based semi-supervised binary classification method that extends the Fisher subspace estimation method by means of a kernel graph covariance measure~\cite{novel}. 
SGT\cite{lv2023revisiting} adopts a static graph structure and Transformer. After adopting PointNet, it confirms the fixed static graph.
Li et al. proposed a graph convolutional architecture TGNet which improves its scale invariance by learning deep features in multiple scale neighborhoods~\cite{tgnet}. 
Chen et al. proposed GAPointNet~\cite{chen2021gapointnet}, which employs self-attention and neighboring-attention mechanism to simultaneously consider self-geometric information and local correlation to  the corresponding neighbours. GAPointNet uses multiple single headed GAP-Layers to learn deep features, and to some extent, each single-head GAP-Layer creates a fixed graph structure, which differs in design and concept from dynamic graphs mentioned in methods such as DGCNN\cite{dgcnn}. ur method aims to re-establish the graph structure after each layer's feature update, so that subsequent feature updates integrate the features transmitted from the previous layer.
The method proposed by Ren et al\cite{dynamic} focuses on the geometry of points and only considers the use of self-attention in the local neighborhood of points, ignoring the feature update of global attention in point clouds. GTNet considers both local and global Transformers simultaneously.
To reasonably utilize the fine-grained information of the point cloud and construct a dynamic graph structure, KCNet~\cite{kcnet} defined the kernel as a group of learnable points, and obtained the geometric affinities from the adjacent points. 
Liu et al. proposed DPAM Module~\cite{dpam} for point agglomeration, compared with aggregation on fixed points, dynamic point aggregation can be more robustly handle all kinds of point cloud data. 
To improve the robustness of point clouds to rotational transformations, ClusterNet~\cite{clusternet} uses hierarchical clustering to learn the features of point clouds in a hierarchical tree. 
Chen et al.
Wei et al. proposed AGConv~\cite{agconv}, which breaks down the barriers of fixed/isotropic kernels and provides a more flexible convolutional approach.
Similar to AGConv~\cite{agconv}, Chen et al. proposed a dual-graph attention convolution network (DGACN)~\cite{dual}.It employs two types of attention convolutions: graph geometric attention convolution (GGAC) and graph embedding attention convolution (GEAC), which obtain geometric attention based on the low-level coordinate space and high-level feature space, respectively.
The method\cite{yun2019graph} proposed by Yun et al. have naming similarities with GTNet. This method is designed for node classification and link prediction, which is fundamentally different from our network designs for point cloud shape classification, part segmentation, and semantic segmentation.
UPU-DGTNet\cite{deng2022upu} proposed by Deng et al uses EdgeConv and MLPs for local feature acquisition, and uses global and local features for Q and K respectively when updating global features. Our method adopts cross-attention and self-attention at local and global levels, respectively.
In DGCNN~\cite{dgcnn}, each layer in the network uses EdgeConv to obtain the local geometric representation, and the dynamic update process of the feature map captures similar semantic features over long distances. 
However, in the local neighborhood, DGCNN sets the maximum value of the neighboring features as the features of centroids, and only the neighbors with the largest feature values affect the centroid features, thus the weak edge-association neighbors have no effect on the centroid features.
GTNet adopts an attention within the local neighborhood to update the center point in a weighted sum of all neighboring point features.

\textbf{Point Cloud Transformers.} Recently, Transformer has become very popular in the field of point cloud.
Many researchers have designed supervised methods~\cite{pan20213d,park2022fast,mao2021voxel,he2022voxel} based on Transformer.
Point Cloud Transformer~(PCT)~\cite{pct} is the first local feature extraction module that uses an intra-domain self-attention mechanism to obtain centroid features. 
Point Transformer~\cite{pt} applies the self-attention mechanism to the local range of each point, and embeds the positional encoding in the input.
It only adopts attention locally, while our method chooses to apply attention globally as well.
SST~\cite{fan2022embracing} and SWFormer~\cite{sun2022swformer} learned about Swin Transformer~\cite{swin}, using Bird's Eye View (projected through point clouds) as input.
PatchFormer~\cite{patchformer} solves the problem of high-computational cost of Point Transformer by estimating a set of patches as bases in the point clouds and replaces the key vector with bases, which reduces the complexity from O($N^2$) to O($MN$), where $N$ is the number of original input points and M is the number of bases, M$\ll$N. 
Cloud Transformer~\cite{cloudtf} combines spatial Transformers with translation, rotation and scaling invariance, and adds 2D/3D mesh features to address the shortcomings of Transformer's poor timeliness, which greatly improves the model efficiency. 
PVT~\cite{pvt} mainly consists of a voxel branch and a point branch, the voxel branch obtains coarse-grained local features by running Sparse Window Attention, and the point branch extracts fine-grained global features by performing Relative Attention or External Attention. 
In addition, some recent Transformer-based models adopt self-supervised learning (SSL) strategy to learn generic and useful point cloud representations from unlabeled data~\cite{voxelmae,mvjar,masked}.
Chen et al. proposed the Masked Voxel Jigsaw and Reconstruction~(MV-JAR)~\cite{mvjar}, which adopts a Reversed-Furthest-Voxel-Sampling strategy to solve the uneven distribution of LiDAR points. Voxel-MAE~\cite{voxelmae}~is a simple masked autoencoding pre-training scheme. This model uses a Transformer-based 3D object detector as the pre-trained backbone to process voxel. SSL methods avoid the need for extensive manual annotations, but with lower performances of the results. Supervised models~\cite{pvt,cloudtf,pct} require labeled data to train and can achieve higher results of the test.

\section{Method}
\label{sec:Method}

\subsection{Backbone}
\label{sec:intro_model}
In this paper, we exploit the advantages of graph-based and Transformer-based methods to design a deep learning model named GTNet, which learns local fine-grained and global coarse-grained features on input to enhance the feature representation, thereby improving the performances of classification and segmentation. Similar to DGCNN~\cite{dgcnn}, we take the geometric coordinate information of the point clouds as input, then we adopt the Coordinate Adjustment Network~(composed of three convolutions, three linear layers, five Batch Normalization, and five activation function layers) to enhance the invariance of rotation and translation. Next we use the feature extraction network to learn the deep representation of points, and finally uses multiple stacked MLPs~(mainly composed of four convolutions, three Batch Normalization, and three activation function layers) to predict the segmentation result. The implementation of Graph Transformer is shown in Figure \ref{fig:architecture}.

\begin{figure}[htbp]
	\centering
 \setlength{\abovecaptionskip}{0.cm} 
\setlength{\abovecaptionskip}{0.cm} 
\setlength{\belowdisplayskip}{3pt} 	
	\includegraphics[width=0.5\textwidth]{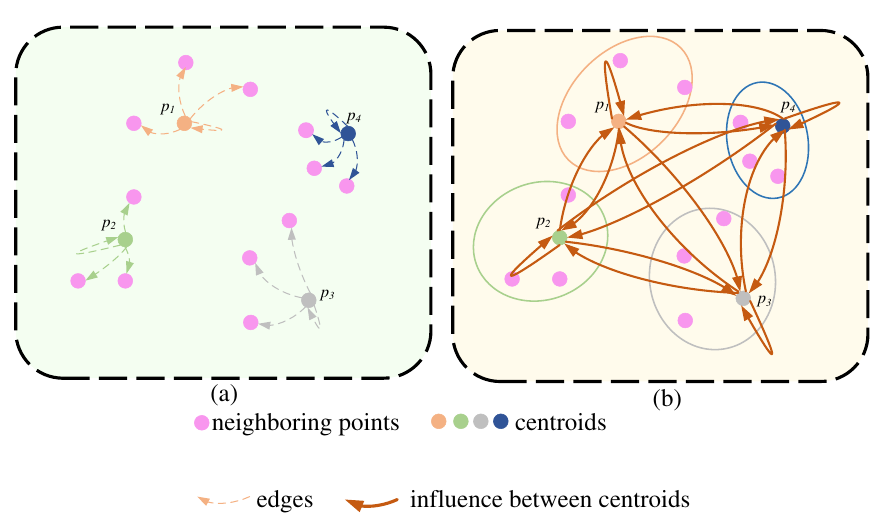}
	\caption{Two types of attention in local Transformer and global Transformer. Figure (a) shows centroids adopt Local Transformer to generate local fine-grained feature in centroids' neighborhood, the connection between each centroids and their neighboring points is considered as edges. The input of the Global Transformer in Figure (b) is the local features of the centroid after the aggregation of the neighborhood features, and the global features of a centroid rely on all centroids.}
	\label{fig:local&global}
\end{figure}

Our feature extraction block Graph Transformer consists of two modules: 
1) Local Transformer, which generates a feature graph using feature dependency between point cloud inputs, and then uses the intra-domain cross-attention mechanism to perform weighted summation of features for different neighboring points which have edge relations, thus to generate new centroid features and set them as the input of Global Transformer; 
2) Global Transformer, using the self-attention mechanism in the global context, which further enhances the contextual information of the features. These two types of attention is shown in Figure \ref{fig:local&global}.
The differences between Local Transformer and Global Transformer are as follows:
1) The scope of calculations differs, with Local Transformer performing on local neighboring points and Global Transformer processing on all centroids;
2) The attention mechanism used is distinct. As Local Transformer adopts intra-domain cross-attention mechanism, and Global Transformer uses self-attention mechanism. In addition, these two types of mechanism differ in handling Query and Key vectors. 

Next we introduce each module of Graph Transformer in the bottom-up form. In Section \ref{sec:Feature_map_design} we describe the implementation of feature graph for Local Transformer; in Section \ref{sec:Graph_Transformer} we introduce the fundamental framework of Graph Transformer: Local Transformer and Global Transformer; in Section \ref{sec:GTNet} we detail the GTNet network for part segmentation and the update process of dynamic graph.

\begin{figure}[htbp]
	\centering
 \setlength{\abovecaptionskip}{0.cm} 
\setlength{\abovecaptionskip}{0.cm} 
\setlength{\belowdisplayskip}{3pt} 	

	\includegraphics[width=0.5\textwidth]{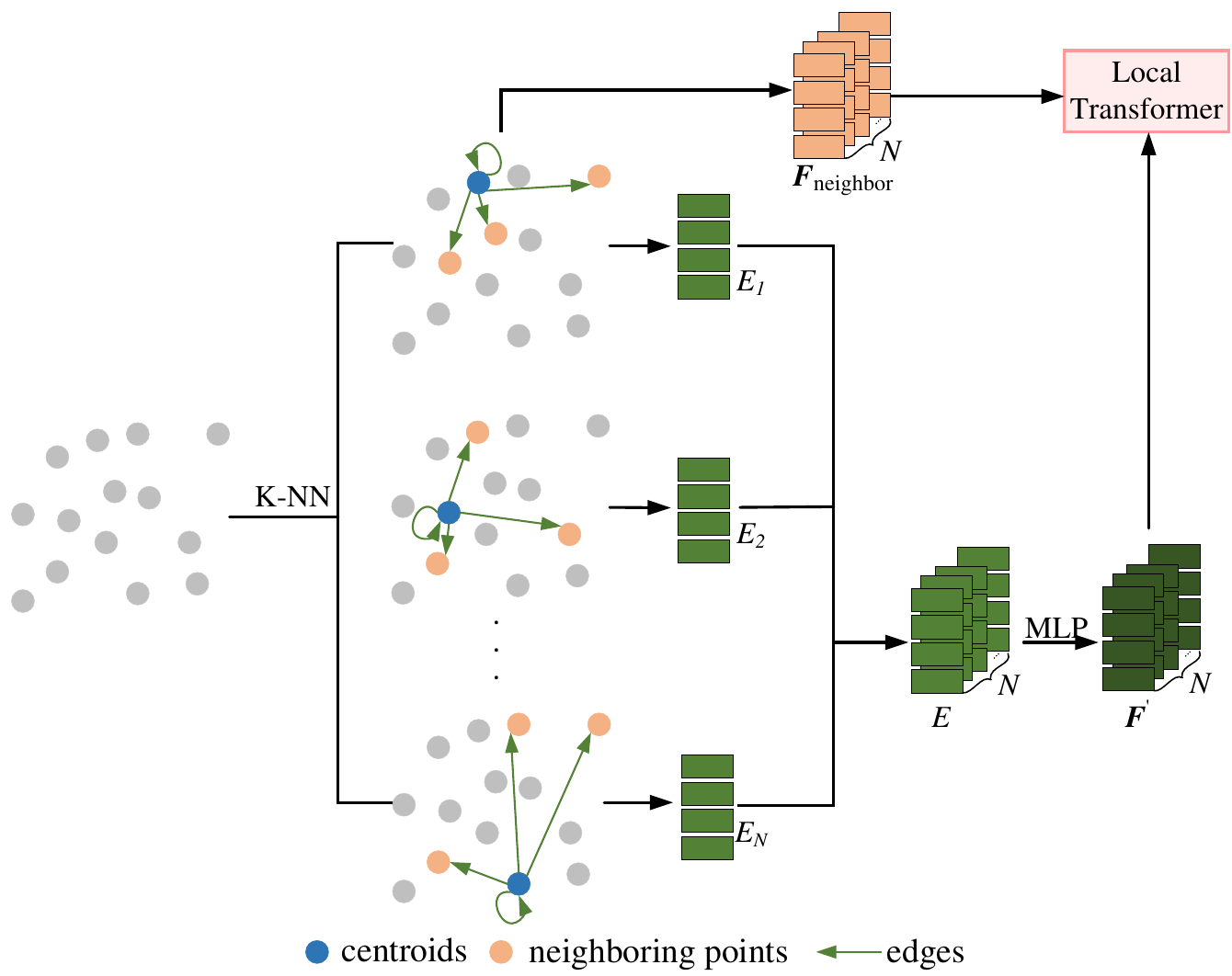}
	\caption{The process of graph generation and feature encoding. We regard all points as centroids, perform K-NN on all centroids in their respective neighborhood, set $K$ to 4, and finally obtain ${\boldsymbol F}_{n e i g h b o r}$ composed of neighboring point features and $\boldsymbol{E}$ composed of edge features.}
	\label{fig:encoding}
\end{figure}

In our model, we need to construct graph structures on the input point clouds for Graph Transformer in Section~\ref{sec:Graph_Transformer}. As shown in Figure \ref{fig:encoding}, before each Graph Transformer, we'll establish the graph relation. Based on this graph relation, we then update the centroids' neighborhood features ${\boldsymbol F}_{n e i g h b o r}$ and edge relations $\boldsymbol{E}$ for Graph Transformer.

\subsection{Feature graph generation}
\label{sec:Feature_map_design}
\textbf{Input data.} It is assumed that the point cloud $\boldsymbol{P}=\left\{p_1, p_2, \ldots, p_N\right\}$ contains $N$ points, and its corresponding feature $\boldsymbol{F}=\left\{f_1, f_2, \ldots, f_N\right\}$ is the input for creating the graph structure. We respectively represent the centroids and centroid features mentioned below as $\boldsymbol{P}$ and $\boldsymbol{F}$.

\textbf{Establishment of graph.} Unlike the learning of independent points in PointNet~\cite{pointnet}, we select each point $\left\{p_1, p_2, \ldots, p_N\right\}$ from the point set $\boldsymbol{P}$ as centroids, then we obtain the set of the $K$ nearest neighbors of the centroids through the spatial coordinate information or the learning features, we'll discuss updating the dynamic graph through the coordinate space and feature spaces in Section~\ref{sec:GTNet} and Figure \ref{fig:update}. Too small value of $K$ makes point clouds in dense areas obtain too little effective information, and too large value of $K$ causes the point clouds in sparse areas introduce too much noise, see Section~\ref{sec:abl_studies} for the discussion on the choice of $K$. 
We use the $i$th centroid $p_i$ and its neighbor $U\left(p_{i} , K\right)=\left\{p_{i 1}, p_{i 2}, \ldots, p_{i K}\right\}$ to construct the graph, and denote the graph as $G_{i}=\{p_{i},E_{i}\}$, where $E_{i}=\left\{e_{i j}\vert j=1,2,\dots,K\right\}$ represents the edge relations between the centroid and its neighboring points. We use $G=\{G_{1},G_{2},\ldots,G_{N}\}$ to represent the graph generated by all the centroids and their corresponding neighboring points. Due to the uneven distribution of points, the neighborhood of different centroids may overlap partially or not overlap at all, so $e_{i j}$ and $e_{j i}$ may not exist simultaneously in the graph.

\textbf{Feature encoding.} The edge relations $\boldsymbol{E}=\{E_{1},E_{2},\ldots,E_{N}\}$ have two types of representations depending on how the neighborhood is acquired.
The first representation of $e_{i j}$ can be expressed as following:
\begin{equation}\label{eq:eij_1}	
	e_{i j}=\psi(f_{j})=w_{j}\cdot f_{j}
\end{equation}
where $\psi$ is the MLP operation~(composed of a layer of linear and an activation function layer), $w_{j}$ is the learned feature weight, $f_{j}$ is the feature information of $p_{j}$, $p_{j}$ is a neighboring point of centroid $p_{i}$, ``$\cdot$" denotes the dot product operation. 
This representation only considers the absolute features of neighboring points.
The edge relation $e_{i j}$ is only associated with the features $f_{j}$ of neighboring point $p_{j}$ and does not relate to the features $f_{i}$ of the centroids $p_{i}$. This neglects the irregular geometric space of the point clouds, leading to a lack of shape perception and contextual information for $e_{i j}$. To associate $p_{i}$ and $p_{j}$ in $e_{i j}$, we express the edge $e_{i j}$ with the second representation as follows:
\begin{equation}\label{eq:eij_2}	
	e_{i j}={\delta}{\big(}f_{i j}{\big)}=w_{i j}\cdot concat{\big(}{\big(}f_{j}-f_{i}{\big)},f_{i}{\big)} 
\end{equation}
where $\delta$ mainly composed of two linear layers and an activation function layer. $w_{ij}$ is the shared weight and $f_{i}$ is the features of $p_{i}$, $f_{ij}$ is associated both with $f_{i}$ and $f_{j}$. In this representation of $e_{i j}$, we concatenate $f_{j}-f_{i}$ and centroid features $f_{i}$ to enhance the perception of local shapes. In Section~\ref{sec:abl_studies}, we'll discuss the performances of our model with or without feature encoding.

When the graph structures are constructed, next is to proceed with our Graph Transformer in Section~\ref{sec:Graph_Transformer}.

\subsection{Graph Transformer}
\label{sec:Graph_Transformer}
Before conducting Graph Transformer to extract features, we use the graph generation method mentioned in Section~\ref{sec:Feature_map_design} to obtain the centroids' neighborhood and their corresponding edge relations $G=\{G_{1},G_{2},\ldots,G_{N}\}$.
As shown in Figure \ref{fig:architecture}, our feature extraction block Graph Transformer consists of Local Transformer and Global Transformer. These two parts are described in detail below.

\begin{figure}[htbp]
	\centering
 \setlength{\abovecaptionskip}{0.cm} 
\setlength{\abovecaptionskip}{0.cm} 
\setlength{\belowdisplayskip}{3pt} 	
	\includegraphics[width=0.5\textwidth]{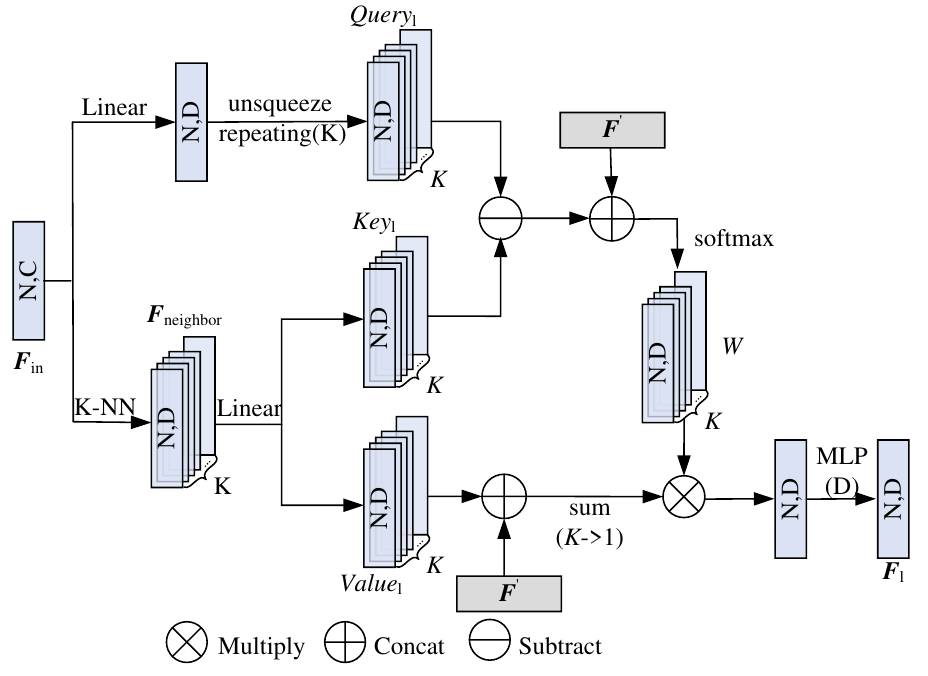}
	\caption{Structure of Local Transformer. Local Transformer firstly uses the dynamic graph to obtain the neighboring points by K-NN, and then conduct weighted summation of features for different neighboring points which are with edge relations. ${\boldsymbol F}^{'}$ is the feature encoding generated by the edge relations $\boldsymbol{E}$, which enhance the perception of local shapes, $K$ is the number of neighbor points, $C$ is the dimension of the input features, and $D$ is the dimension of the generated features.}
	\label{fig:Local_Transformer}
\end{figure}

\textbf{Local Transformer.} 
As shown in Figure \ref{fig:Local_Transformer}, based on the constructed graph relations $U\left(p_{i} , K\right)=\left\{p_{i 1}, p_{i 2}, \ldots, p_{i K}\right\}$, we firstly calculate $Query_{l}$,~$Key_{l}$ and $Value_{l}$ vectors on the centroid features ${\boldsymbol F}_{i n}\subseteq\mathbb{R}^{N\times C}$ of $p_{i}$ and the corresponding neighborhood features ${\boldsymbol F}_{n e i g h b o r}\subseteq\mathbb{R}^{N\times K\times C}$ of $\left\{p_{i 1}, p_{i 2}, \ldots, p_{i K}\right\}$:
\begin{equation}\label{eq:qkv_local}
	\left\{\begin{array}{l l}{{Q u e r y_{l}=\boldsymbol{F}_{i n}\cdot w_{ql}
		}}\\ {{(K e y_{l},V a l u e_{l})=\boldsymbol{F}_{n e i g h b o r}\cdot\left(w_{kl},w_{vl}\right)}}\end{array}\right. 
\end{equation}
where $Query_{l}\subseteq\mathbb{R}^{N\times D}$, $Key_{l}$, $V a l u e_{l}\subseteq\mathbb{R}^{N\times K\times D}$, $w_{ql}$, $w_{kl}$, $w_{vl}\subseteq\mathbb{R}^{C\times D}$, and $D$ is the feature dimensions after mapping.

To map the dimension of $Query_{l}$ from $\mathbb{R}^{N\times D}$ to $\mathbb{R}^{N\times K\times D}$, we perform the following operation:
\begin{equation}\label{eq:queryl_1}
	Query_{l}^{'}=\gamma(Query_{l}) 
\end{equation}
where $\gamma$ is the unsqueeze function, $Query_{l}^{'}\subseteq\mathbb{R}^{N\times K\times D}$. We learned vector attention in Point Transformer~\cite{pt} and subtracted query vector and key vector. We then calculate the weight matrix $W$ with $Query_{l}^{'}$ and $Key_{l}$ to let each neighboring point constrain the centroid features (neighbors with stronger relations hold more weight, and neighbors with more dissimilarity own less weight): 
\begin{equation}\label{eq:W}
	W=Q u e r y_{l}^{'}-K e y_{l}+\boldsymbol{F}^{'} 
\end{equation}
where $\boldsymbol{F}^{'}$ is the deep feature after encoding. Inspired by the position encoding proposed by Zhao et al. in Point Transformer~\cite{zhao2021point}, we use the edge relations generated by Eq.~\eqref{eq:eij_2} 6in Section \ref{sec:Feature_map_design} as the feature encoding to enhance the perception of local shapes (as shown in Figure \ref{fig:encoding}):
\begin{equation}\label{eq:deep_feature}
	\boldsymbol{F}^{'}=\tau\left({\boldsymbol{E}}\right) 
\end{equation}
where the edge relations $\boldsymbol{E}$ are the shallow features, $ \tau $ is the MLP operation~(composed of two linear and a nonlinear activation function. layer).

After acquiring the deep feature $ \boldsymbol{F}^{'} $, we further learn the new features and perform the aggregation function to obtain the local features $ \boldsymbol{F}_{l} $:
\begin{equation}\label{eq:fl}
	\boldsymbol{F}_{l}=\Lambda\left(W^{\prime}\cdot\left(V a l u e_{l}+\boldsymbol{F}^{'}\right)\right) 
\end{equation}
where $ \Lambda $ is the aggregation function which uses max-pooling or avg-pooling for the neighborhoods to obtain local fine-grained features (see the aggregation analysis in Section \ref{sec:abl_studies} for a discussion of the two aggregation functions), $W^{'}$ is the updated weight:
\begin{equation}\label{eq:w1}
	W^{'}=s o f t m a x\left(\frac{W}{\sqrt{d_{kl}}}\right)
\end{equation}
where $ \sqrt{d_{kl}} $ is the scaling factor, the normalization of $W$ is adopted to  speed up the convergence of the model.

\begin{figure}[htbp]
	\centering
 \setlength{\abovecaptionskip}{0.cm} 
\setlength{\abovecaptionskip}{0.cm} 
\setlength{\belowdisplayskip}{3pt} 	
	\includegraphics[width=0.5\textwidth]{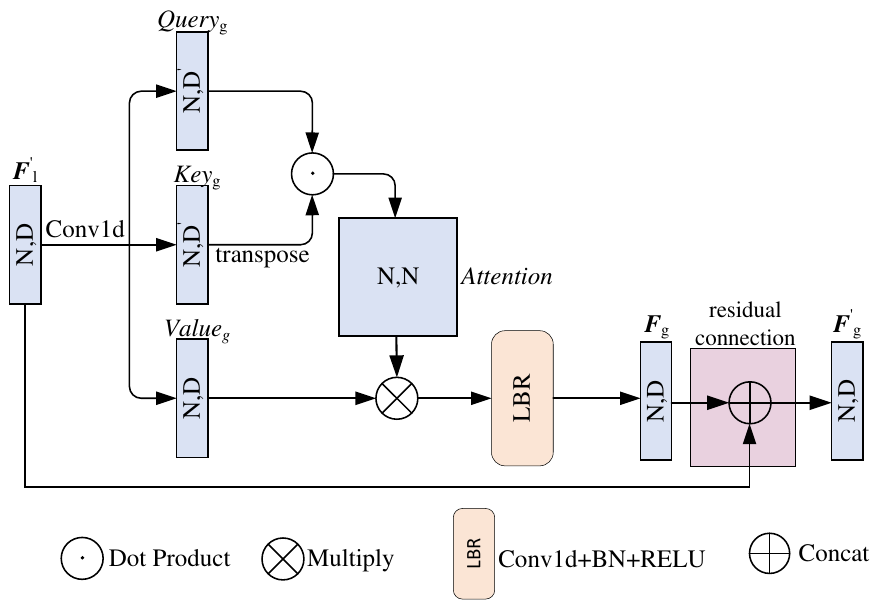}
	\caption{Structure of Global Transformer. It uses a global self-attention mechanism, where the feature generation of each centroid is derived from all the centroids of the input, which enhances the global representation of the features, and uses residual connection to alleviate overfitting and gradient disappearance problems during the training period. $Attention$ is the generated weight matrix, and $LBR$ is the feature alignment layer.}
	\label{fig:Global_Transformer}
\end{figure}
\textbf{Global Transformer.} 
The process of implementing the Global Transformer is shown in Figure \ref{fig:Global_Transformer}. We firstly compute the $ Qu e r y_{g} $, the $ Key_{g} $ and $  Value_{g} $ vectors with the input features $ \boldsymbol{F}_{l}^{'} $ ($ \boldsymbol{F}_{l}^{'}=\boldsymbol{F}_{in}+\boldsymbol{F}_{l}$):
\begin{equation}\label{eq:qkv_global}
	{\left(Qu e r y_{g},{K}e y_{l},V a l u e_{l}\right)=\boldsymbol{F}_{l}^{'}\cdot\left({w}_{qg},{w}_{kg},{w}_{vg}\right)} 
\end{equation}
where  $ Qu e r y_{g} $,  $ Key_{g}\subseteq\mathbb{R}^{D\times D^{'}} $, $  Value_{g}\subseteq{\mathbb{R}^{D\times D}} $, ${w}_{qg}$, ${w}_{kg}\subseteq\mathbb{R}^{D\times D^{'}}$, ${{ w}_{vg}\subseteq{\mathbb{R}^{D\times D}}}$, $ D^{'}=D/4 $. Then we use the self-attention mechanism to obtain the global features $ \boldsymbol{F}_{g} $: 
\begin{equation}\label{eq:fg}
	\boldsymbol{F}_{g}=softmax\left({Q u e r y_{g}}\cdot{K e y_{g}^\top}\right)\cdot V a l u e_{g} 
\end{equation}
where $\cdot$ is dot product.

Inspired by the residual connection, we apply it to update the global features from $\boldsymbol{F}_{g} $ to $ {\boldsymbol{F}}_{g}^{'}$, which can suppress the overfitting of the model and avoid the problem of gradient disappearance and degradation:
\begin{equation}
	{\boldsymbol{F}}_{g}^{'}=\boldsymbol{F}_{l}^{'}+\xi\big(\boldsymbol{F}_{l}^{'}-\boldsymbol{F}_{g}\big) 
\end{equation}
where $ \xi $ is the feature alignment layer which includes convolutional layer, normalization layer and non-linear activation function layer.

\subsection{GTNet and dynamic graph update process}
\label{sec:GTNet}

\begin{algorithm}[H]
	\caption{$ \boldsymbol{F}_{shape} $ Gathering Algorithm} 
\label{alg:alg1}
	\begin{algorithmic}[1]
 \Require Point clouds $\boldsymbol{P}$, label $L$, neighbor$ \_ $num $K$, feature$ \_ $block$ \_ $num $M$
 \State$output=\left[~\right]$
		\State $\boldsymbol{F}_{in}=\boldsymbol{P}$
		\For{$m=1$ to $M$}
		\State $ \boldsymbol{F}_{g}^{'}=Graph$ \_ $Transformer_{m}\left(\boldsymbol{F}_{in},K\right) $
		\State $output.append\left(\boldsymbol{F}_{g}^{'}\right)$
		\State $ \boldsymbol{F}_{in}= \boldsymbol{F}_{g}^{'}$
		\EndFor	
		\State $ \boldsymbol{F}_{agg}=maxpool\left(concat\left(output\right)\right)$	
		\State $ \boldsymbol{F}_{shape}=concat\left(\boldsymbol{F}_{agg},MLP\left(L\right)\right)$
	\end{algorithmic}
\end{algorithm}

In this paper, we can observe in Figure \ref{fig:architecture} that the input of each Graph Transformer block uses the output of the previous Graph Transformer block. 
We create the feature graph for each Graph Transformer block by performing K-NN on the output features of the previous Graph Transformer block.

As shown in Figure \ref{fig:update}, the iteration of the Graph Transformer can be regarded as the learning process of the dynamic graph, and we can also obtain deeper feature from these iteration processes. For the part segmentation task, GTNet also introduced label information $L\in\mathbb{R}^{1\times k}$ and directly generated the updated label features, where $ k $ is the number of categories contained in the dataset.

As shown in Algorithm \ref{alg:alg1}, after performing the $ M $-layer Graph Transformer, we concatenate the output features of each Graph Transformer, then obtain the learned feature $ \boldsymbol{F}_{agg} $ after the max-pooling. Finally, we concatenate $ \boldsymbol{F}_{agg} $ with the label information, and obtain the final output feature $ \boldsymbol{F}_{shape} $.

\section{Experiment}
\label{sec:exp}
To verify the performances of GTNet, we conduct experiments on the ModelNet40, ShapeNet Part, and S3DIS datasets to implement shape classification, part segmentation and large scene semantic segmentation. Our experiments use PyTorch to implement GTNet, and the model is trained on NVIDIA GeForce RTX 3080Ti GPU.

\begin{table}
\centering
	\renewcommand{\arraystretch}{1}
	\caption{Results of the shape classification task on the ModelNet40 dataset. Compared with the Transformer-based methods in the table, the performance improvement of GTNet may be due to the adoption of attention on local neighborhood with dynamic graphs and the introduction of global attention; Compared with the graph-based methods in the table, dynamic graphs perform better than static graphs, and our method uses weighted sums on neighborhood features as the aggregation form; Compared with the method that combines Graph and Transformer simultaneously, GTNet does not adopt downsampling processing, which retains more original input features, and we propose to design dynamic graphs on the feature space. In the Type column, (DG) represents the dynamic graph, and (SG) represents the static graph. It should be noted that in the table, we have retrained and tested the results of PointNet++ and PCT}
\label{table:modelnet40}
    \scalebox{0.7}{
	\begin{tabular}{c|c|cc}
		\hline 
		Method    &Type    & OA(\%)            & mAcc(\%)          \\ \hline
        A-SCN\cite{xie2018attentional} &Transformer-based &89.8 &87.4 \\
        P2SResLNet\cite{wu2024point} &- &90.6 &89.2 \\
        LO-Net\cite{li2024classification} &- &91.2 &88.9 \\
        PAT\cite{yang2019modeling} &Graph-based(SG) \& Transformer-based &91.7 &- \\
		PointNet++\cite{pointnet++} &-   & 92.5          & 89.7             \\
  3D-GCN\cite{lin2020convolution} &Graph-based(SG)   & 92.1          & -             \\
		DGCNN\cite{dgcnn}    &Graph-based(DG)    & 92.2          & 90.2          \\
         GAPointNet\cite{chen2021gapointnet} & Graph-based (SG) \& Transformer-based &92.4 &89.7 \\
         DDGCN\cite{ddgcn} &Graph-based(DG) &92.7 &90.4 \\
		PointASNL\cite{pointasnl}  &Transformer-based   & 92.9          & -             \\
           
		PCT\cite{pct}    &Transformer-based       & 93.2 & 90.0\\
   DGCNN+MMI\cite{improving}   &Graph-based(DG)  & 93.2          & 90.6 \\
   FatNet\cite{kaul2021fatnet} &Transformer-based &93.2          & 90.6    \\ 
   ACNN\cite{hassan2023residual} &- &93.2 &91.2 \\

   APES\cite{apes} &Transformer-based &\textbf{93.8} &- \\
   \hline
		Ours     & Graph-based(DG) \& Transformer-based    & 93.2 & \textbf{92.6}  \\ \hline
		
	\end{tabular}}
\end{table}

\subsection{Shape classification on the ModelNet40 dataset}\label{sec:modelnet40} 

\textbf{Data and metrics.} ModelNet40\footnote[1]{https://shapenet.cs.stanford.edu/media/modelnet40\_ply\_hdf5\_2048.zip} dataset contains 12311 shapes in 40 different categories, of which 9843 shapes are used for training and 2468 shapes are used for testing. In the experiment, we sample 1024 points uniformly from each model and utilize their coordinate information as input. Overall accuracy (OA) and category accuracy (mAcc) are adopted as the evaluation metrics of models:
\begin{equation}\label{eq:OAmAcc}
	\left\{\begin{array}{l l}{OA=\frac{\sum_{i=1}^{k}R_{i}}{\sum_{i=1}^{k}N_{i}}}\\ {{\mathrm{}}}\\ {{m A c c=\frac{\sum_{i=1}^{k}\frac{R_{i}}{N_{i}}}{k}}}\end{array}\right. 
\end{equation}
where $ R_{i} $ represents the number of correctly predicted points in category $i$, and $N_{i}$ denotes the actual number of points belonging to category $i$.


\textbf{Results.} The results in Table \ref{table:modelnet40} show that GTNet achieves the highest values in both OA and mAcc, which prove that GTNet is more capable in shape classification than most other models based on Graph and Transformer. 
Compared with graph-based model DGCNN~\cite{dgcnn}, GTNet improve 1\% on OA and 2.4\% on mAcc. DGCNN also adopts the dynamic graph structure, which demonstrating that the model with dynamic graph combined with Transformer (GTNet) can show better classification ability than the model with dynamic graph combined with convolution (DGCNN). Compared with Transformer-based model PointASNL~\cite{pointasnl},GTNet improve 0.3\% on OA. Compared to Method GAPointNet~\cite{chen2021gapointnet}, which also combines Graph and Transformer, GTNet improve 0.8\% on OA and 2.9\% on mAcc. GTNet exceeds some particularly classic deep learning models, that has an improvement of OA than PointNet~\cite{pointnet} and PointNet++~\cite{pointnet++} by 4\% and 1.3\% respectively, and also owns a 2.4\% improvement over mAcc than the second highest results in the table, which demonstrating that GTNet can perform better feature learning and achieve higher accuracy in different categories. We believe that the reason why GTNet has better performance is because it applies the principle of dynamic graphs and uses Transformers to obtain deeper feature both locally and globally.

\begin{table*}
	\centering
	\renewcommand{\arraystretch}{1.1}
	\caption{Results of the part segmentation task on the ShapeNet Part dataset. Compared to graph-based} methods, adding Transformer may result in performance improvement; Compared to Transformer-based models, GTNet includes Global Transformer; Compared to the method of combining both graph and Transformer, we have incorporated the principle of dynamic graph. In the Type column of the table, GB represents graph-based methods, TB represents Transformer--based methods, and G\&TB represents both graph and Transformer based methods.
	\label{table:shapenet}
	\resizebox{2\columnwidth}{!}{
 \scalebox{1}{
		\begin{tabular}{c|c|c|cccccccccccccccc}
			\hline 
			Method   &Type  & mIoU(\%)          & Airplane      & Bag           & Cap           & Car           & Chair         & Earphone      & Guitar        & Knife         & Lamp          & Laptop        & Motorbike     & Mug           & Pistol        & Rocket        & Skateboard    & Table         \\ \hline
             A-SCN\cite{xie2018attentional} &TB &84.6 &83.8 &80.8 &83.5 &79.3 &90.5 &69.8 &91.7 &86.5 &82.9 &96.0 &69.2 &93.8 &82.5 &62.9 &74.4 &80.8 \\
             GAPointNet\cite{chen2021gapointnet}  &G\&TB 
                & 84.7 &84.2 &\textbf{84.1} &\textbf{88.8} &78.1 &90.7 &70.1 
                &91.0 &87.3 &83.1 &\textbf{96.2} &65.9 &95.0 &81.7 &60.7 &74.9 &80.8 \\
			PointNet++\cite{pointnet++} &-  & 85.1          & 82.4          & 79.0          & 87.7          & 77.3          & 90.8          & 71.8          & 91.0          & 85.9          & 83.7          & 95.3          & 71.6 & 94.1          & 81.3          & 58.7          & 76.4          & 82.6          \\
            3D-GCN\cite{lin2020convolution} &GB &85.1 &83.1 &84.0 &86.6 &77.5 &90.3 &74.1 &90.9 &86.4 &83.8 &95.6 &66.8 &94.8 &81.3 &59.6 &75.7 &82.6 \\
			DGCNN\cite{dgcnn}  &GB    & 85.2          & 84.0          & 83.4 & 86.7          & 77.8 & 90.6          & 74.7          & 91.2 & 87.5 & 82.8          & 95.7 & 66.3 & 94.9 & 81.1 & \textbf{63.5} & 74.5 & 82.6          \\
            PMMNet\cite{cao2024pmmnet} &- &85.2 &- &- &- &- &- &- &- &- &- &- &- &- &- &- &- &- \\
            FatNet\cite{kaul2021fatnet} &TB &85.5 &- &- &- &- &- &- &- &- &- &- &- &- &- &-  &- &- \\
            APES\cite{apes} &TB &\textbf{85.8} &- &- &- &- &- &- &- &- &- &- &- &- &- &- &- &- \\
			\hline
			Ours   &G\&TB    & 85.5          & 84.1          & 80.3          & 81.5          & 78.2        & 90.9          & 70.2          & 91.6 & 87.5        & 84.8          & 95.8 & 61.2          & 93.9          & 83.3          & 53.6          & 75.4          & \textbf{83.1}          \\ \hline
			
	\end{tabular} }}
\end{table*}

\subsection{Part segmentation on ShapeNet Part dataset}
\label{sec:shapenet}

\textbf{Data and metrics.} ShapeNet Part\footnote[2]{https://shapenet.cs.stanford.edu/media/shapenet\_part\_seg\_hdf5\_data.zip} dataset contains 16881 3D shapes which belong to 16 different categories, each category contains 2-5 parts, and all the categories are subdivided into a total of 50 different parts. In this experiment, we uniformly sample 2048 points for each shape, and use their coordinate information as input. The experimental results are finally evaluated by mIoU:
\begin{equation}\label{eq:miou}	
mIoU=\frac{\sum_{i=1}^{s}IoU_{i}}{s} 
\end{equation}
where $s$ represents the number of 3D shapes in the dataset, $IoU_{i}$ represents the Intersection-over-Union of the $i$-th shape.


\begin{figure}[htbp]
	\centering 
 \setlength{\abovecaptionskip}{0.cm} 
\setlength{\abovecaptionskip}{0.cm} 
\setlength{\belowdisplayskip}{3pt} 	
	\includegraphics[width=0.5\textwidth]{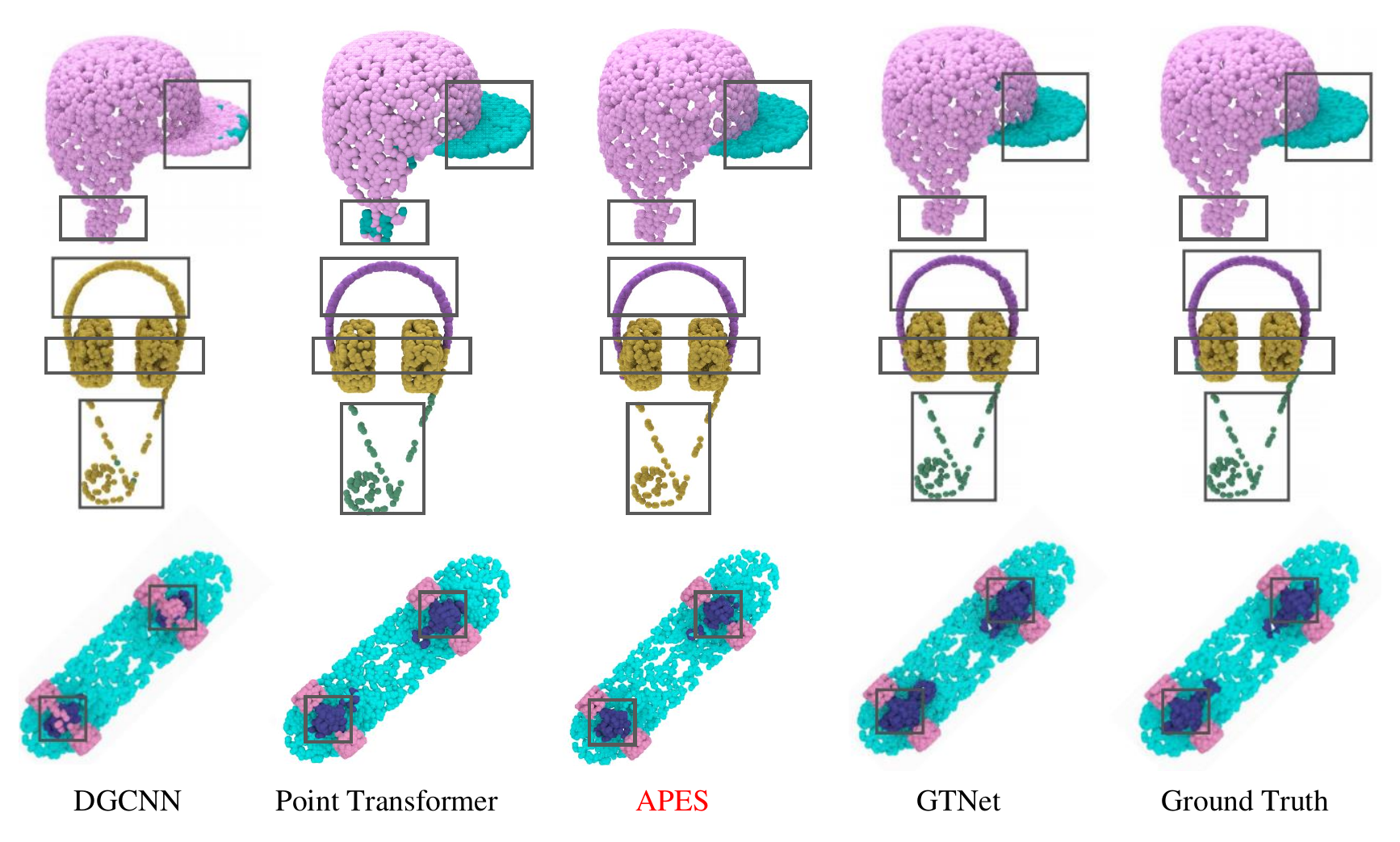}
	\caption{Part of visualization results for part segmentation in ShapeNet Part. For each set, from left to right: DGCNN, Point Transformer, APES, GTNet, and ground truth. The object's segmentation performance can be visually observed in the figure's boxes, and this distinction may be attributed to GTNet's utilization of Transformers at both local and global levels.}
	\label{fig:shapenet_red}
\end{figure}

\textbf{Results.} Table \ref{table:shapenet} shows the performance of GTNet compared with other models on the ShapeNet Part dataset. We calculate the mean of IoU for all shapes in each category and the mean of IoU for all tested shapes (mIoU) respectively. Similar to the shape classification experiment, we compared the performance of GTNet with PointNet~\cite{pointnet} and PointNet++~\cite{pointnet++}. Our model achieves 1.8\% improvement on mIoU compared to PointNet. Compared to PointNet++, we also improve results on several categories (1.7\% on the Airplane, 1.1\% on the Lamp, etc.). Compared to most other methods~\cite{xie2018attentional,chen2021gapointnet,lin2020convolution,dgcnn,kaul2021fatnet} based on Graph or Transformer, GTNet achieved many higher performances, such as 90.9\% on the Chair, 83.3\% on the Pistol, 84.8\% on the Lamp, etc. In addition, we also visualized the part segmentation results of DGCNN, Point Transformer~\cite{pt}, APES~\cite{apes}, and GTNet in Figure \ref{fig:shapenet_red}. In this table, we found that GTNet does not always perform best on each category. The reason is that multi-scale has not been implemented, and a unified $K$ value is used in each category when obtaining local features. The composition structure of different objects is different, such as categories such as Laptop, Guitar, Table, etc., which have fewer components, so the noise doped by neighboring points within the $K$ value range is less; However, categories such as Motorbike and Rocket have a larger number of components, and some of these categories contain fewer points, making them more prone to doping with noise compared to categories with fewer components. Using smaller neighborhoods is more suitable for categories with more components.

\subsection{Large indoor scene semantic segmentation on S3DIS}
\label{sec:s3dis}

\textbf{Data and metrics.} S3DIS\footnote[3]{https://cvg-data.inf.ethz.ch/s3dis/Stanford3dDataset\_v1.2\_Aligned\_Version.zip} dataset contains point cloud data in 6 indoor areas, consisting of 272 rooms. There are 13 semantic categories in the scenes: bookcase, chair, ceiling, beam and others. In this experiment, each room is scaled to a 1m × 1m cell block, in each block, we sample 4096 points for training, and use all points of the block for testing. We adopt 6-fold-cross-validation and OA to evaluate the performance.

\begin{figure*}[htbp]
	\centering
 \setlength{\abovecaptionskip}{0.cm} 
\setlength{\abovecaptionskip}{0.cm} 
\setlength{\belowdisplayskip}{3pt} 	
	\includegraphics[width=0.9\textwidth]{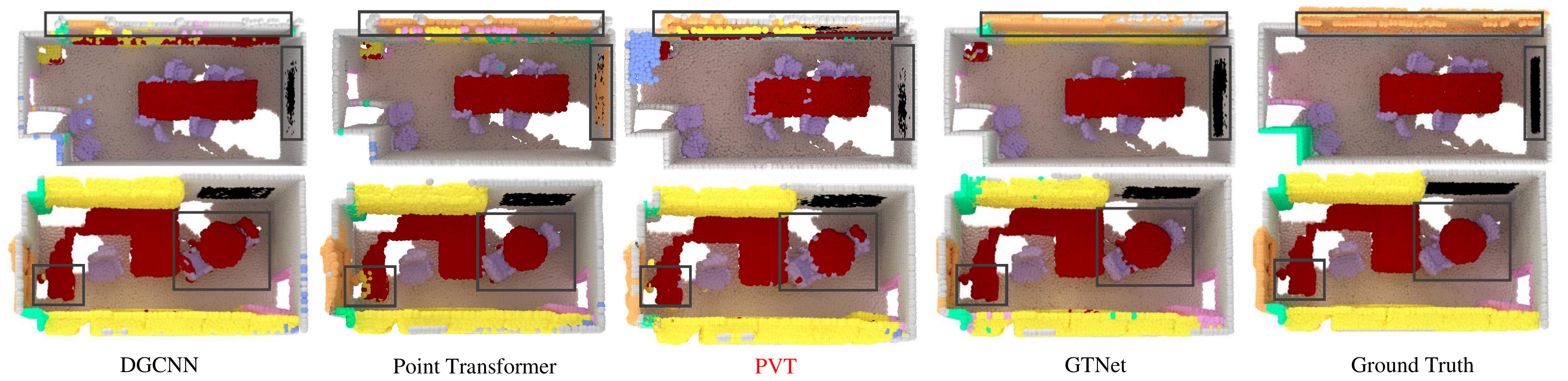}
	\caption{Part of visualization results for large indoor scene semantic segmentation in S3DIS dataset. From left to right: DGCNN, Point Transformer, PVT, GTNet, and ground truth. The visual gap in indoor scenes can be seen from the boxes.}
	\label{fig:s3dis_red}
\end{figure*}

\textbf{Results.} As shown in Table \ref{table:s3dis}, comparing with the existing state-of-the-art models such as PointNet~\cite{pointnet}, A-SCN~\cite{xie2018attentional}, DGCNN~\cite{dgcnn}, SPGraph~\cite{SPGraph}, and PAT~\cite{yang2019modeling}, GTNet significantly outperforms most of them in 6-fold-cross-validation. Compared with DGCNN~\cite{dgcnn}, GTNet improves 2.5\% on OA and 8.2\% on mIoU, demonstrating that the combination of intra-domain cross-attention mechanism and global self-attention mechanism enables the model to acquire richer contextual information in the feature learning process. We also visually compared our model with the results of DGCNN, Point Transformer~\cite{pt}, and PVT~\cite{pvt} in Figure \ref{fig:s3dis_red}. Compared with A-SCN~\cite{xie2018attentional}, GTNet improves 5\% on OA and 11.5\% on mIoU. The performance improvement may come from the fact that in our method, the input of Global Transformer is the detailed features updated by Local Transformer. At the same time, we found that in large scene data, segmentation performance is better than object data, possibly because the large scene data contains relatively more points, and thus the influence of the number of neighboring points $k$ is weaker.

\begin{table}
	\centering
	\renewcommand{\arraystretch}{1.1}
	\caption{Semantic segmentation results on S3DIS. GTNet performs better than the Graph or Transformer based methods in the table, potentially because it utilizes dynamic graphs locally and applies Transformers simultaneously at both local and global areas.}
	\label{table:s3dis}
 \scalebox{0.7}{
	\begin{tabular}{c|c|cc}
		\hline 
		Method    &Type  & OA(\%)            & mIoU(\%)           \\ 
		\hline			
            A-SCN\cite{xie2018attentional} &Transformer-based &81.6 &52.8 \\
		
		DGCNN\cite{dgcnn}    &Graph-based   & 84.1          & 56.1           \\ 
		SPGraph\cite{SPGraph}  &Graph-based   & 85.5          & 62.1           \\ 
        HAPGN\cite{chen2020hapgn} &Graph-based \& Transformer-based  &85.8 &62.9\\
        PAT\cite{yang2019modeling} &Graph-based \& Transformer-based &- &\textbf{64.3} \\
  \hline
		Ours     &Graph-based \& Transformer-based   & \textbf{86.6}          & \textbf{64.3}           \\
		\hline
	\end{tabular} }
\end{table}

\begin{table}
	\centering
	\renewcommand{\arraystretch}{1.1}
	\caption{Space and time complexity on different tasks.}
	\label{table:complexity}
	\scalebox{0.7}{
	\begin{tabular}{c|c|cc}
		\hline 
		Task   &Method   & FLOPs/sample  & Params \\
        \hline
        \multirow{3}{*}{Shape classification}
        &PointNeXt-S\cite{qian2022pointnext} &1.6G &1.4M \\
        &RepSurf-U\cite{ran2022surface} &2.4G  &6.8M \\
        &Ours  &4.3G &2.1M \\ 
        \hline
        \multirow{3}{*}{Part segmentation}
        &PT$^2$\cite{zhao2021point} &17.1G &9.14M \\
        &PointNeXt-S\cite{qian2022pointnext} &110.2G &22.5M \\
        &Ours &4.6G &1.8M \\ 
        \hline
        \multirow{3}{*}{Semantic segmentation} 
        &RepSurf-U\cite{ran2022surface} &1.0G &0.9M \\
        &PointNeXt-XL\cite{qian2022pointnext} &84.8G  &41.6M \\
        &Ours &8.421G &1.696M \\ 
        \hline
	\end{tabular}
    }
\end{table}

\subsection{Model Complexity}
Table \ref{table:complexity} presents a comparison of the space and time complexity for shape classification, part segmentation, and semantic segmentation tasks between GTNet and other existing models. Below is an explanation of the complexity experiment of our model: in the shape classification task, FLOPs is 8.607G and params is 2.194M; in the part segmentation task, FLOPs is 9.340G and params is 1.796M; in the semantic segmentation task, FLOPs is 16.842G and params is 1.696M.

\subsection{Ablation studies.}
\label{sec:abl_studies}
In this section, we perform various ablation studies on the ShapeNet Part dataset to verify the effectiveness of different modules in GTNet. In the following experiment, we demonstrated that the use of a backbone network (feature extraction block consisting of four layers of Graph Transformer, where Graph Transformer updates features locally and globally using local-cross attention and global self-attention, local neighborhood setting $K$ to 20, max-pooling for aggregation, feature encoding in Transformer, and residual connections in feature extraction layer) can achieve the best performance.

\textbf{Transformer analysis.} The main parts of Graph Transformer are Local Transformer and Global Transformer. In the ablation learning, we remove one of them while remain the other one to assess the individual effectiveness of these Transformer components. As shown in Table \ref{table:Component_analysis}, model A only uses Local Transformer, Model B directly uses Global Transformer after applying 1×1 convolution to the input to align the dimensions, and Model C keeps both Local Transformer and Global Transformer. Model B and C are two different baselines, and Model A has been improved based on both baselines. With the removal of Local Transformer or Global Transformer, the mIoU will only decrease by 1.58\% and 1.96\% respectively, which shows that these two components can learn the deep feature of the point clouds even if they perform separately. All the results show that the combination of these two components is better than taking a single one.

\begin{table}
	\centering
	\renewcommand{\arraystretch}{1.1}
	\caption{Ablation study of the numbers of feature extraction layer.}
	\label{table:number}\scalebox{0.7}{
	\begin{tabular}{c|ccc}
		\hline 
		Number  & OA(\%)           & mAcc(\%)          & mIoU(\%)           \\ \hline
		2   & 93.81 & 81.83 &84.20 \\
		3                 &93.94          &82.31          &84.65          \\
		4                 & \textbf{94.12} & \textbf{83.63} & \textbf{85.14}      
		\\ \hline
	\end{tabular} }
\end{table}

\textbf{Numbers of feature extraction layer.} This experiment investigates the number of feature extraction layers in Graph Transformer, which determines the feature update performance of the model. The results are shown in Table \ref{table:number}, and the best performance is achieved when the number is set to 4. When the number is small ($Number=2$ or $Number=3$), GTNet does not extract enough deep features to represent point features. As the number of layers increases, the number of parameters in the model will continue to increase, resulting in higher requirements for computational costs. Currently, our equipment only supports running models with $Number=4$.

Composition of edge features $\boldsymbol{e_{ij}}$. This experiment investigates the selection of $f_j$ and $f_{ij}$ in $e_{ij}$, which determines the composition of point neighborhood features. The results are shown in Table \ref{table:fij}, where the best performance is achieved when the neighborhood feature uses $f_{ij}=(concat (f_j-f_i), f_i)$. When using $f_j$ for features, only local neighboring point features are considered, without considering the original features of the center point. When using $f_{ij}=concat (f_j, f_i)$ for features, the feature differences between neighboring and central points are not considered. When the feature adopts $f_{ij}=(concat (f_j+f_i), f_i)$, adding too many center point features (approximately twice) increases redundant information and weakens the effect of neighboring point features.

\begin{table}
	\centering
	\renewcommand{\arraystretch}{1.1}
	\caption{Ablation study of $f_j$ and $f_{ij}$ in dynamic graph edges $e_{ij}$.}
	\label{table:fij}\scalebox{0.7}{
	\begin{tabular}{c|c|ccc}
		\hline 
		$Features$ &Design  & OA(\%)           & mAcc(\%)           & mIoU(\%)           \\ \hline
		$f_{j}$    &$f_j$    & 93.93 & 82.63 &84.56 \\
		$f_{ij}$    &concat$(f_j,f_i)$                 &93.95          &83.78         &84.67          \\
  $f_{ij}$    &concat$((f_j+f_i),f_i)$   &93.91          &82.36         &84.53          \\
		   $f_{ij}$    &concat$((f_j-f_i),f_i)$          & \textbf{94.12} & \textbf{83.63} & \textbf{85.14}       
		\\ \hline
	\end{tabular} }
\end{table}

\begin{table}
	\centering
	\renewcommand{\arraystretch}{1.1}
	\caption{Ablation study of Local Transformer and Global Transformer, ``\checkmark" indicates the adoption of this module, we identify ``LT" as the Local Transformer, and ``GT" as the Global Transformer.}
	\label{table:Component_analysis}\scalebox{0.7}{
	\begin{tabular}{c|cc|ccc}
		\hline 
		Model & LT & GT & OA(\%)           & mAcc(\%)           & mIoU(\%)           \\ \hline
		A     & \checkmark                 & \checkmark                  & \textbf{94.12} & \textbf{83.63} & \textbf{85.14} \\
		B     & \checkmark                 & ~                  & 93.36          & 77.66          & 83.18          \\
		C     & ~                 & \checkmark                  & 93.42          & 80.67          & 83.56          
		\\ \hline
	\end{tabular} }
\end{table}

\textbf{Aggregation analysis.} In Algorithm \ref{alg:alg1} of Section \ref{sec:GTNet}, the feature extraction network comprises multiple Graph Transformers, and we concatenate the output of each Graph Transformer to aggregate the features. Here, we adopt four aggregations: max, avg, add (max, avg) and concat (max, avg) to perform the ablation test, where max is the max pooling and avg is the average pooling, add (max, avg) is to directly add the results of max pooling and average pooling, and concat (max, avg) is to concatenate the results of max pooling and average pooling. From the results shown in Table \ref{table:Aggregation_operation.}, we can observe that only taking max pooling is better than only taking average pooling, for the two operations combining max and avg, the concatenating improves the mIoU by 0.53\% compared with direct adding, single using max as the aggregation operation is able to extract more representative features in the feature update process.
\begin{table}
	\centering
	\renewcommand{\arraystretch}{1.1}
	\caption{Ablation study of aggregations. The performance was tested using four aggregations: max, avg, add (max, avg) and concat (max, avg).}
	\label{table:Aggregation_operation.} \scalebox{0.7}{
	\begin{tabular}{c|ccc}
		\hline 
		Function          & OA(\%)             & mAcc(\%)           & mIoU(\%)           \\ \hline
		max+avg           & 93.62          & 81.44          & 84.01          \\
		concat (max, avg) & 93.85          & 81.08          & 84.54          \\
		avg               & 93.86          & 81.76          & 84.44          \\
		max               & \textbf{94.12} & \textbf{83.63} & \textbf{85.14} \\ \hline
	\end{tabular} }
\end{table}

\textbf{Number $K$ of neighboring points.} This experiment analyzes the number of neighbors set in Local Transformer, which determines the neighborhood range of the centroids. The results are shown in Table \ref{table:k_value}, the best performance is achieved when $K$ is set to 20. GTNet could not extract sufficient contextual information for model prediction when the neighborhood range is small ($K=5$ or $K=10$ or $K=15$). The implementation of the intra-domain cross-attention mechanism may introduce too many noise points when the neighborhood range is large ($K=25$), and this also directly leads to a decrease in the accuracy of the model.
\begin{table}
	\centering
	\renewcommand{\arraystretch}{1.1}
	\caption{Ablation study for the number $K$ of neighboring points on local neighborhoods.}
	\label{table:k_value} \scalebox{0.7}{
	\begin{tabular}{c|ccc}
		\hline $K$
		& OA(\%)             & mAcc(\%)           & mIoU(\%)            \\ 
		\hline
		5  & 93.49          & 79.45          & 83.74           \\
		10 & 93.83          & 80.04          & 84.33           \\
		15 & 93.79          & 81.47          & 84.43           \\
		20 & \textbf{94.12} & \textbf{83.63} & \textbf{85.14}  \\
		25 & 93.57          & 80.07          & 83.79           \\
		\hline
	\end{tabular} }
\end{table}

\textbf{Feature encoding.} Local Transformer takes feature encoding to enhance the perception of local shapes. In this investigation, we test its effect by taking and removing feature encoding $ \boldsymbol{F}^{'} $. The results are shown in Table \ref{table:encoding_feature}. In the table, remove the model with feature encoding as the baseline. If the feature encoding is missing, the performance of the model decreases significantly by 1.48\%, which also reflects that the feature encoding proposed in this paper is practical and can improve the performance of the model.

\begin{table}
	\centering
	\renewcommand{\arraystretch}{1.1}
	\caption{Ablation study of feature encoding, A indicates the model without feature encoding $\boldsymbol{F}^{'} $, and B represents the model with feature encoding $\boldsymbol{F}^{'} $.}
	\label{table:encoding_feature}\scalebox{0.7}{
	\begin{tabular}{c|ccc}
		\hline 
		Model & OA(\%)             & mAcc(\%)           & mIoU(\%)           \\ \hline
		A      & 93.56          & 80.49          & 83.66          \\
		B     & \textbf{94.12} & \textbf{83.63} & \textbf{85.14} \\
		\hline
	\end{tabular} }
\end{table}

\textbf{Residual connection.} Global Transformer uses the residual connection for the output of the self-attention mechanism. To demonstrate that the residual connection can enhance the learning ability of the model, we test the models with and without the residual connection respectively. The results are shown in Table \ref{table:residual_connection}. In the table, model without residual connection is used as baselines. The model with the residual connection improves OA by 0.15\%, mAcc by 1.57\%, and mIoU by 0.37\% comparing to the model without residual connection, which proves that the residual connection can enhance the learning ability of our model.
\begin{table}
	\centering
	\renewcommand{\arraystretch}{1.1}
	\caption{Ablation study of residual connection, A indicates the model without residual connection, B represents the model with residual connection.}
	\label{table:residual_connection} \scalebox{0.7}{
	\begin{tabular}{c|ccc}
		\hline 
		Model         & OA(\%)             & mAcc(\%)           & mIoU(\%)           \\ \hline
		A & 93.97\textbf{} & 82.06\textbf{} & 84.77\textbf{} \\
		B & \textbf{94.12} & \textbf{83.63} & \textbf{85.14} \\
		\hline
	\end{tabular} }
\end{table}

\subsection{Implementation details}\label{sec:implement_Detail}

The feature extraction network consists of four Graph Transformer blocks. Throughout the entire experiment, we uniformly use $(C, D)$ to denote the pre-defined parameters of each Graph Transformer block, where $C$ and $D$ are the dimensional of the input and output features respectively. 

In shape classification task, the feature extraction network uses a four-layer stacked Graph Transformer, and the input and output dimension of the four blocks are set to (3, 64), (64, 64), (64, 128), and (128, 256) respectively, with the increasing dimensions to learn more finer-grained information. 
During the training process, our model sets the learning rate of SGD optimizer to 0.0001, batch$ \_ $size to 8, and iteratively learns for 250 epochs.

In part segmentation task, the number of feature extraction block layers is the same as in shape classification task, and the input and output dimensions of the four blocks are set to (3, 96), (96, 96), (96, 96), and (96, 96) respectively. In the training process, our model is set with a batch size of 10 and trained for 200 epochs. We use an SGD optimizer with a learning rate of 0.01, in which the momentum size is 0.9 and the weight decay is 0.0001. The neighborhood size $K$ of K-NN is set to $20$.

In large indoor scene semantic segmentation task, the feature extraction network settings of the model are consistent with part segmentation. During training, the neighborhood size $K$ of K-NN is set to $15$, and the model is set with a batch size of 4 and trained for 50 epochs.

\section{Conclusion}
\label{summary}
In this paper, we design the deep learning model GTNet for various tasks of point clouds. GTNet is mainly composed of Graph Transformer blocks and MLPs. 
Graph Transformer uses the dynamic graph and Transformer to learn features in the local and global patterns, where Local Transformer is adopted to extract fine-grained features with all neighboring points, and Global Transformer is used to obtain coarse-grained features. 
In addition to using coordinates to generate graphs, our method uses the output features of each Graph Transformer to continuously update the graph relations dynamically.
We also introduce the feature encoding in the local feature learning to enhance the perception of local shapes, and conduct residual connection in GTNet to enhance the learning ability of our model.

In future work, we want to design models not only more efficiently, but also multi-scale (each layer combines multiple different sizes of neighbors). In this paper, we only design the model on shape classification, part segmentation and semantic segmentation tasks, and have not extended it to other domains, we also want to study the application in point cloud registration, 3D reconstruction and other fields.

\section*{Acknowledgments}

This work is supported by Shaanxi Province Key Research and Development Projects 2020KW-068, General Project of Education Department of Shaanxi Provincial Government under Grant 22JK058, Xi'an Key Laboratory of Aircraft Optical Imaging and Measurement Technology Open Fund Project 2023-006, National Natural Science Foundation of China under Grant 52367015, China Postdoctoral Science Foundation under Grant 2024M750897, and Jiangxi Provincial Natural Science Foundation under Grants 20224BAB204051 and 20232BAB214064.

\bibliographystyle{cag-num-names}
\bibliography{refs}



\end{document}